\documentclass[11pt]{article}
\usepackage{amssymb}
\usepackage{amsthm}
\usepackage{amsmath,amsfonts,amssymb}
\usepackage[ruled,linesnumbered,lined,commentsnumbered]{algorithm2e}
\usepackage[top=1.5in, bottom=1.5in, left=1.25in, right=1.25in]{geometry}

\usepackage[mathscr]{eucal}
\usepackage{exscale}
\usepackage[numbers]{natbib}
\usepackage{bm}
\usepackage{eqlist}
\usepackage[dvipsnames]{color}
\usepackage[Lenny]{fncychap}
\usepackage{multirow}
\usepackage{graphicx}
\usepackage{hyperref}
\usepackage{framed}
\usepackage{longtable}
\usepackage{multirow}
\usepackage[toc,page]{appendix}

\usepackage{tabu}

\font\bigbf=cmbx10 scaled \magstep3

\begin{document}

\title{\bigbf Deep-MAPS: Machine Learning based\\ Mobile Air Pollution Sensing}

\author{Jun Song$\,^a$
 \quad Ke Han$\,^{b,a,*}$
  \\\\
 \textit{$^a$Department of Civil and Environmental Engineering, Imperial College London, UK}\\
\textit{$^b$School of Transportation and Logistics, Southwest Jiaotong University, China}\\\\
$^*$Corresponding author: kxh323@gmail.com 
\\
}

\maketitle

\begin{abstract}
Mobile and ubiquitous sensing of urban air quality has received increased attention as an economically and operationally viable means to survey atmospheric environment with high spatial-temporal resolution. This paper proposes a machine learning based mobile air pollution sensing framework, called {\it Deep-MAPS}, and demonstrates its scientific and financial values in the following aspects. (1) Based on a network of fixed and mobile air quality sensors, we perform spatial inference of PM2.5 concentrations in Beijing (3,025 km$^2$, 19 Jun-16 Jul 2018) for a spatial-temporal resolution of 1km$\times$1km and 1 hour, with over 85\% accuracy. (2) We leverage urban big data to generate insights regarding the potential cause of pollution, which facilitates evidence-based sustainable urban management. (3) To achieve such spatial-temporal coverage and accuracy, Deep-MAPS can save up to 90\% hardware investment, compared with ubiquitous sensing that relies primarily on fixed sensors. 
\end{abstract}


\section{INTRODUCTION}\label{iinntt}
\vspace{-0.1in}
Severe deterioration of urban {\it air quality} (AQ) around the globe, especially in developing countries due to aggressive urbanization and motorization, has posed barriers to economic development and major threats to public health. Health Effects Institute \cite{HEI2018} estimates that 95\% of the world's population is exposed to air quality considered unsafe by the World Health Organization (WHO). WHO further estimates that 3.7 million deaths in 2012 were caused by outdoor air pollution, nearly 90\% of which were in developing countries. In China, up to 1.3 million deaths per year are due to air pollution, and the monetary value based on death and illness is \$1.4 trillion in 2010 and rising. In the United Kingdom, an estimated 40,000 people per annum die prematurely due to air pollution.

Detailed information about air pollution, including its sources and dynamics on a city scale, is of critical importance to public health and sustainable urban management \citep{LLFLL2015}. For decades, fixed-location AQ monitoring stations have provided AQ information for various purposes including information dissemination, AQ inference/prediction, and policy appraisal. In recently years, mobile air quality sensing has received increased attention as an economically and operationally viable means to survey urban atmospheric environment  \citep{CK2003, White2012}. Compared to fixed monitoring stations, the network of mobile sensors can offer high-resolution and dense coverage of urban atmospheric environment at comparatively low costs \citep{Kumar2015}. A number of initiatives have taken place world wide to collect AQ information based on mobile sensing and communication technologies, as reported by \cite{Wallace2009, Wang2009, Mead2013, MZT2018, Alvear2018}.

The global market for ubiquitous AQ sensing is rapidly expanding with a 14.3\% compounded average growth rate, reaching \$530m by 2024 \citep{IR2019}. In China,  public funding allocated to air pollution mitigation has reached 25bn RMB (3.6bn USD) in 2019, with a 25\% increase over 2018 \citep{Xinhua2019}. Given the major investment in hardware, however, the efficacy of mobile sensing remains rather limited, as information available is restricted to sensor measurements without effective mining of data essential to decision making. In this paper, we demonstrate the scientific and financial values of Deep-MAPS, a machine learning framework in \underline{m}obile \underline{a}ir \underline{p}ollution \underline{s}ensing, in three main ways: (1) It allows air quality information to be inferred at fine granularity (1km$\times$1km, hourly) at city-wide scale, even with sparse fixed and mobile sensing coverage. (2) It leverages urban big data to generate insights regarding the potential cause of pollution, which facilitates evidence-based sustainable urban management. (3) It has the potential to substantially reduce the capital investment of high-density AQ monitoring in urban areas.

\section*{SIGNIFICANCE}
\vspace{-0.1 in}
Detailed and accurate information of air quality is crucial for public health and sustainable urban development. Portable air quality sensors can offer detailed information on urban atmospheric environment. We demonstrate how machine learning (Deep-MAPS) can effectively harness the power of mobile sensing and urban big data to provide city-scale, fine-granular air pollution maps, with 85\% accuracy and under 10\% of the cost of fixed-location sensing. Our findings also offer quantitative insights regarding the relationship between pollutant concentration and urban configuration or dynamics. This work enables critical assessment of public exposure, and offers decision support for evidence-based policy making. Deep-MAPS is a scientifically and financially viable tool for sustainable urban management with a potential for global adaptation.

 \section{RESULTS}
\vspace{-0.1in}

\subsection{Beijing case study}
\vspace{-0.1in}
The study area is a 55km$\times$55km square in Beijing, covering the Sixth Ring Road as shown in Figure \ref{figgridpmcs}. This is divided into 3,025 1km$\times$1km {\it spatial grids}. The spatial inference is carried out for  grid with an hourly resolution in a four-week period between 19 June and 17 July 2018. In this paper, we focus on PM2.5 as it has highly adverse health effects and increases mortality risks to the public under long-term exposure. We aim to infer the hourly PM2.5 concentrations in the study area, with a total number of $3,025\times 28\times 24=2,032,800$.

\subsection{Validation of spatial inference results for PM2.5}
\vspace{-0.1in}
A total of 50,736 labels were obtained from fixed-location and mobile sensing data; see Section \ref{subsecPMmeasure} for more details. A five-fold cross validation, along with RMSE, SMAPE and $R^2$, are used to assess the validity and accuracy of the machine learning model (Deep-MAPS). Table \ref{tabresults} compares Deep-MAPS with several benchmark methods including spatial interpolation (SI), k-nearest neighbors (KNN), and support vector regression (SVR).

\begin{table}[h!]
\centering
\begin{tabular}{c|c|c|c|c}
\hline
                        & Feature      &  RMSE & SMAPE  & R$^2$
\\
\hline
SI (IDW)           & -                 &  17.80  &  21.88\%    &  0.837
\\
\hline
SI (Kriging)        & -                 &   20.80 &  26.57\%    &  0.777
\\
\hline
\multirow{2}{*}{KNN}               & L     &   23.40    & 25.85\%  &  0.720
\\
\multirow{1}{*}{}                   & L+M         &   15.40    & 17.28\%  &  0.879
\\
\hline
\multirow{2}{*}{SVR}                      & L           &  36.49    & 38.37\%  &  0.320
\\
\multirow{1}{*}{}                             & L+M      &  27.47    & 31.36\%  &  0.614
\\
\hline
\multirow{5}{*}{Deep-MAPS}     & L           &  14.37 & 16.35\%   & 0.895
\\
\multirow{1}{*}{}           & L+M           &  13.88 & 15.54\%  & 0.902
\\
\multirow{1}{*}{}                       & N              &   14.71   & 15.84\%  & 0.889
\\
\multirow{1}{*}{}                         & N+M         &  13.98 & 14.85\% & 0.900
\\
\multirow{1}{*}{}                         & L+M+N     &   13.17 & 14.65\% & 0.911
\\
\hline
\end{tabular}
\caption{\small Performance of different methods and features. IDW: inverse distance weighting; L=local features; N=neighboring features; M=macro features.}
\label{tabresults}
\end{table}

The results show that Deep-MAPS, with all combination of urban features (local features, neighboring features and macro features), reaches an accuracy around 85\% with $R^2=0.9$, which outperform other benchmark methods. In addition, comparing `L' with `L+M' (or `N' with `N+M') indicates that the macro features significantly improves the accuracy of KNN, SVR and Deep-MAPS. This means that it is important to consider regional transport of pollutants, which was not done in any of the machine learning models proposed to date. For Deep-MAPS, the neighboring features `N' (or `N+M') outperforms the local features `L' (or `L+M'), suggesting that the spatial and temporal relationships among urban features should not be ignored in the fitting of AQ data. 

\begin{figure}[h!]
\centering
\includegraphics[width=\textwidth]{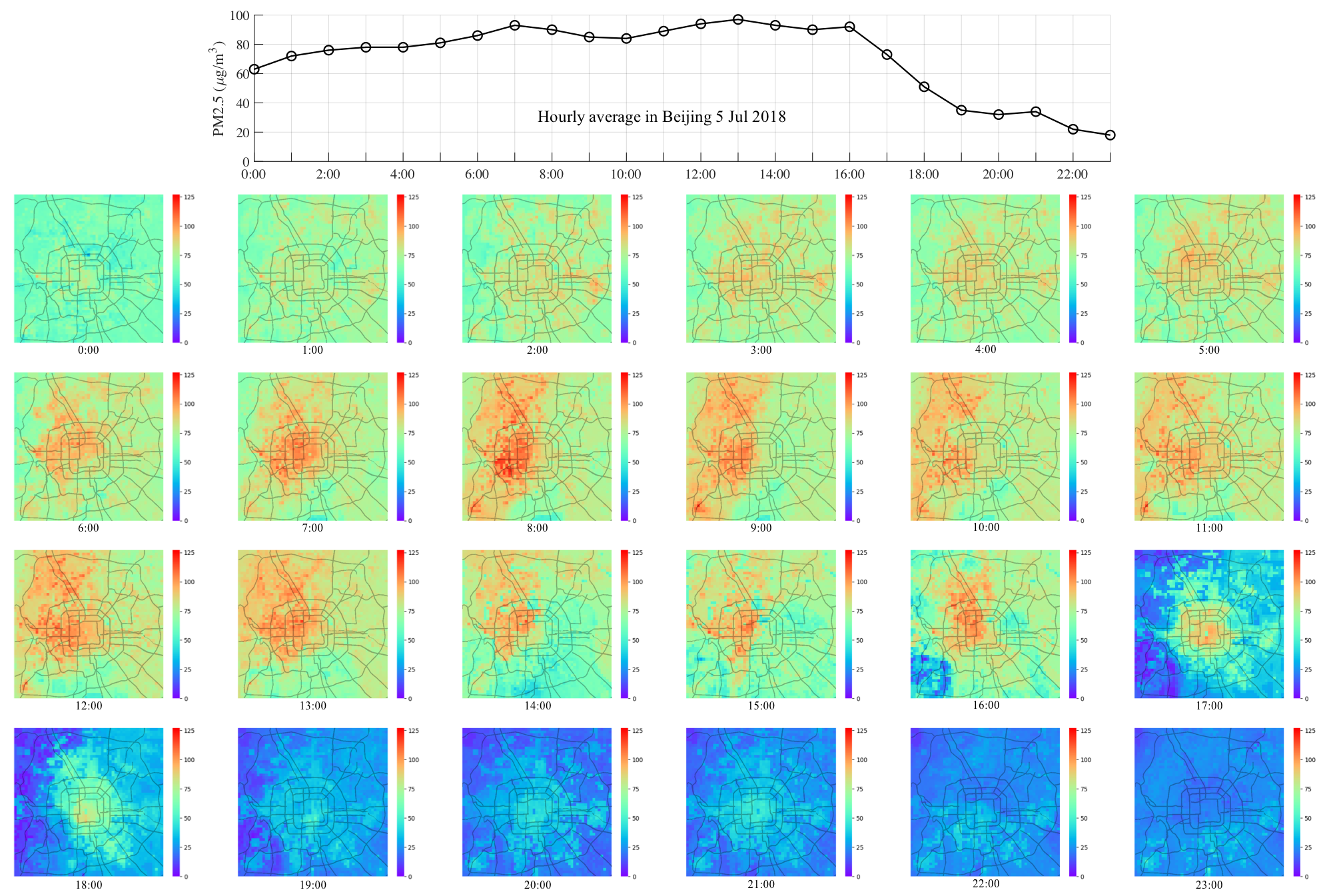}
\caption{\small PM2.5 concentrations ($\mu g/m^3$) in Beijing on 5 July 2018.}
\label{figgridpmcs}
\end{figure}

The result of spatial inference is illustrated for 5 July 2018 in Figure \ref{figgridpmcs}. The hourly PM2.5 concentration maps show relatively high and steady concentrations during most of the day before a sudden drop between 16:00-19:00, which was likely caused by the thundershower in late afternoon. This temporal trend agrees with the PM2.5 concentration in Beijing obtained by averaging fixed-location data, as shown on top of Figure \ref{figgridpmcs}. The fine-grained pollution concentration maps could offer critical insights into population exposure and potential cause of local concentrations.

\subsection{Impact of mobile sensing coverage}
\vspace{-0.1in}
One of the practical concerns of mobile sensing is the amount of mobile coverage required to reach certain level of inference accuracy. To address this problem, we set aside an independent test set of size 9,200 consisting of randomly selected fixed and mobile labels. For the training set, we consider six different cases: all the fixed labels plus $x\%$ of mobile labels, $x=0,\,20,\,40,\,60,\,80,\,100$.

\begin{figure}[h!]
\centering
\includegraphics[width=\textwidth]{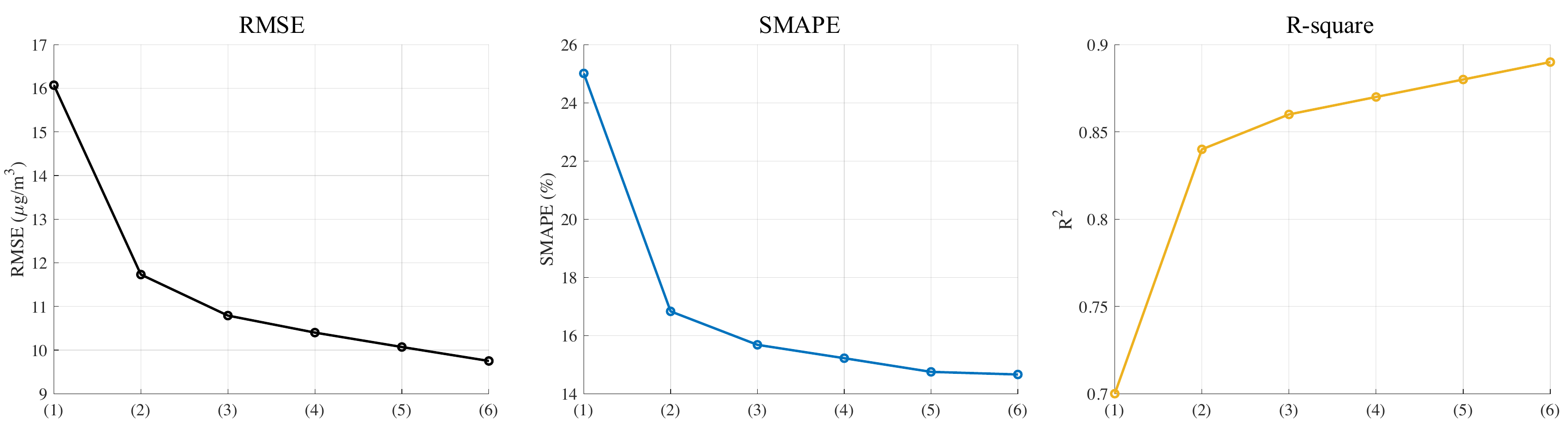}
\caption{\small Performance of Deep-MAPS on an independent test set with different training data. (1): Fixed-location data only; (2): Fixed data + 20\% mobile data; (3): Fixed data + 40\% mobile data; (4): Fixed data + 60\% mobile data; (5): Fixed data + 80\% mobile data; (6): Fixed data + 100\% mobile data.}
\label{figlegs}
\end{figure}

Figure \ref{figlegs} shows the performance of Deep-MAPS with these six training sets. All three performance indices exhibit improving trends as more mobile data points are included in the training set. In particular, the `elbow' point occurs where mobile data first appear in the training set, even in low quantity. Nevertheless, with every increment of mobile data the margin of improvement declines. This is interesting as it suggests that in order to achieve certain level of accuracy, Deep-MAPS only requires limited amount of mobile sensing data. In our case, less than 5\% of the spatial coverage suffices to achieve over 85\% accuracy.

\subsection{Feature analysis and interpretation of the inference results}\label{feaana}
\vspace{-0.1 in}
We analyze the relative importance of different urban features by investigating their weights generated by GBDT, in order to generate insights regarding potential causes and relevant factors of PM2.5 concentrations. The weight of an input feature measures its contribution to the reduction in loss by rendering a better fit between model outputs and observations. 

\begin{figure}[h!]
\centering
\includegraphics[width=\textwidth]{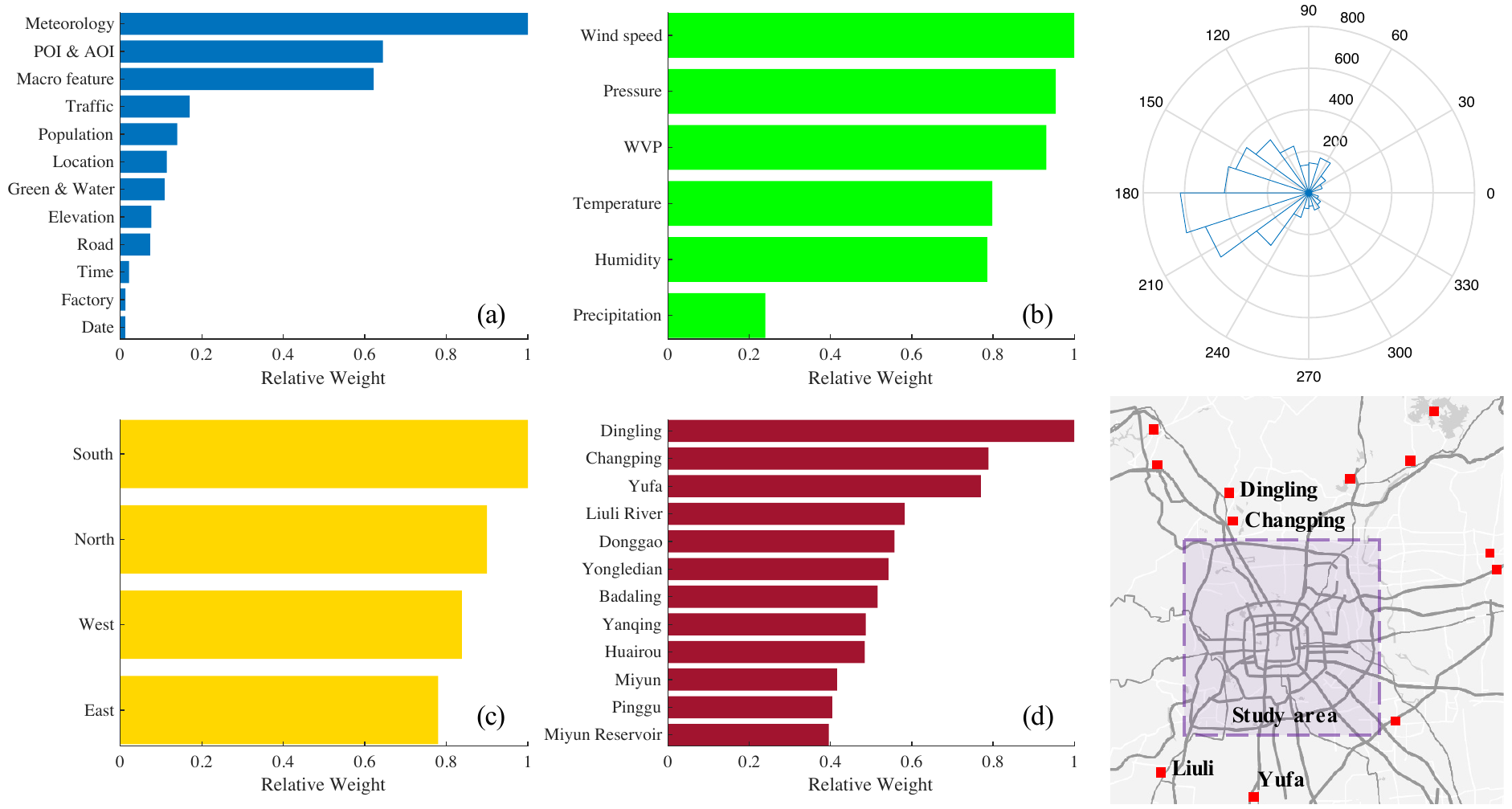}
\caption{\small Feature analysis. (a)-(b) show the relative weights/importance of features internal to the study area. (c)-(d) indicate the regional transport of PM2.5}
\label{figfeatureana}
\end{figure}

Based on results generated from Deep-MAPS , Figure \ref{figfeatureana}(a) shows the relative importance of main feature categories, among which the top five are meteorology, POI\&AOI, macro feature (i.e. regional transport of pollution), traffic conditions and population vitality. Indeed, the concentration of PM2.5 during the summer season is crucially related to meteorological conditions such as wind speed and humidity, the relative weights of which are shown in panel (b). In addition, local restaurants, auto services, residential buildings etc. (represented by POI, AOI) are the main cause of PM2.5 formation; this is followed by traffic which are responsible for fugitive dust and secondary reactions.

Figures \ref{figfeatureana}(c)-(d) are concerned with macro features and the regional transport of PM2.5. The relative weights of individual fixed-location stations outside the study area indicate likely paths of regional transport. In particular, Dingling and Changping, located to the Northwest, have the highest weights, followed by Yufa and Liuli River, which are located to the Southwest. This highlights two main directions of PM2.5 transport (Northwest and Southwest) during 19 June - 16 July 2018. The wind rose diagram in Figure \ref{figfeatureana} provides supportive information regarding the physical transport of PM2.5.

\section{Discussion}
\vspace{-0.1in}

\subsection{Related work}
\vspace{-0.1 in}
In this paper we propose a machine learning framework (Deep-MAPS) to perform spatial inference of PM2.5 concentration on a fine-granular spatial and temporal scale. Conventional `bottom-up' models, such as Computational Fluid Dynamic Models \citep{Chu2005, Parra2010}, Chemical Transport Models \citep{Simpson2012}, and CMAQ model \citep{EPA2019} could offer high model fidelity pertaining to the physical and chemical processes of air pollution. However, they tend to be computationally expensive and rely on assumptions that limit their integration with high-frequency and ubiquitous AQ sensing data.

In contrast to the bottom-up approaches, machine learning aims to infer and predict AQ by learning its spatial and temporal correlation with explanatory factors such as urban layouts, traffic activities, and meteorological conditions. Studies in this line of research consider both fixed-location \citep{ZLH2013, Qi2018} or mobile sensing data \citep{MAM2015, HRBS2017} for air quality inference, but have achieved limited integration of fixed and mobile data in fine-granular spatial inference. Moreover, these models do not offer adequate insights that are important for the mitigation of urban pollution. 

\subsection{Interpretation of urban features}
\vspace{-0.1 in}
A crucial step in achieving the aforementioned AQ inference and managerial insights is the consideration of urban features, which should ideally have a balanced representation of urban configuration and dynamics contributing to the emission and formation of pollutants. In this paper we consider 62 types of urban data, covering geography, land use, transport, public, and meteorology (see Section \ref{subsecUFdata} for details), to capture the spatial and temporal variation of urban features related to air pollution. Furthermore, a novel approach involving macro features is employed to capture the regional transport dynamics outside the study area, which not only improves the inference accuracy but also suggest likely paths of regional transport. 

As we have shown in Section \ref{feaana}, Deep-MAPS allows us to interpret the inference results in terms of the contribution of each input feature. However, this is a preliminary assessment of the correlation between PM concentration and urban features. It should not be misconstrued with causality analysis, i.e. pollution attribution.

\subsection{Potential impact of Deep-MAPS}
\vspace{-0.1 in}

In Aug 2018, the Chinese government launched the Qianliyan Initiative \citep{Qianliyan2018}, which aims to monitor the atmospheric environment (in particular, PM2.5) of the JingJinJi Area (Beijing, Tianjin and Hebei) and 28 surrounding cities. The whole area was divided into 36,793 spatial grids of size 3km$\times$3km, and in major cities like Beijing, the spatial resolution can reach 1km$\times$1km. The capital investment for such a large-scale AQ monitoring project, assuming at least one fixed monitoring station per spatial grid, could be drastically reduced by mobile sensing infrastructure, including Deep-MAPS. Take the study area of this paper for instance. To achieve full coverage of the 3,025 grids, the minimum capital investment is $3,025\times 300$k (RMB/station), which is over 900m RMB, excluding annual maintenance \& admin fees, where 300k RMB is a rather optimistic price estimate for a reliable micro-station. In contrast, Deep-MAPS can achieve 85\% inference accuracy with under 20 mobile and 30 fixed stations, the cost of which is less than 10\% of the 900m investment. This is a considerable saving even with the additional costs of securing and administrating the urban big data required to run Deep-MAPS. Moreover, the benefit of Deep-MAPS goes beyond financial savings by offering additional data and insights as we conveyed in this paper.

\section{Method}
\vspace{-0.1 in}

\subsection{Urban features}\label{subsecUFdata}
\vspace{-0.1 in}
A detailed and balanced representation of urban features is critical for data-driven methods to reduce overfitting and increase the interpretability of the model.  A total of 62 features are grouped into four categories.These features are further distinguished as static and dynamic features, and are aggregated onto the space-time grids for AQ spatial inference. 

\noindent {\bf Geography \& Land Use -} includes land use type (e.g. building, factory, commercial area) and points of interest (POI). 21 categories of POIs are considered, which include restaurants, shopping, schools, hotels, firms, scenic spots etc. Within a 1km$\times$1km spatial grid, the numbers of different types of POIs, as well as the percentages of landmass they occupy (Area of Interest, AOI) are calculated. In addition, the percentages of green and water covers, as well as digital elevation are included to further characterize the geographic features of the grid.

\noindent {\bf Transport -} contains static road network structure and dynamic traffic conditions. Within a 1km$\times$1km spatial grid, the road network information contains number of signalized intersections and the lengths of primary, secondary, tertiary and quaternary roads. These static features are related to traffic volume and fleet composition, as well as the frequency of stop-and-go driving cycles, which are known to contribute to vehicle emissions. The dynamic traffic information is represented by the percentages of roads with light, medium and heavy traffic conditions. Such a categorical feature is a direct indicator of congestion levels, and is related to vehicle speed and emissions. 

\noindent {\bf Public Vitality -} refers to the intensity of social media activities enabled by Location Based Services (LBS). These include Wechat and Sina Weibo log-ins, posts and comments, which are indicators of public vitality in the area. The public vitality data are deemed relevant to instantaneous population density and indirect indicators of land use type (such as business districts and rural areas). The Wechat data is hourly based and the Sina data is aggregated into annual average based on the period Jun 2015 - Jun 2018. 

\noindent {\bf Meteorology -} contains hourly information of local temperature, pressure, water vapor pressure, relative humidity, wind direction and speed, which are measured at 13 meteorological stations (9 within the study area). The hourly meteorological data were used to spatially interpolate grid-based quantities using the inverse distance weighting method.

\subsection{PM2.5 measurements}\label{subsecPMmeasure}
\vspace{-0.1 in}

The 28 fixed monitoring stations within the study area provide hourly concentrations of particulate matters (PM2.5, PM10). The mobile AQ data include PM2.5 and PM10 concentrations as well as temperature and relative humidity, which are collected by mobile sensors mounted onto vehicles. The trajectories of the 15 vehicles that carry the mobile sensors were map-matched, and the mobile PM2.5 measurements were pre-processed (including outlier removal and noise reduction) and aggregated into spatial grids and time intervals, before calibrated using fixed-location data as reference.

The calibration of mobile PM2.5 data is performed using multivariate regression involving time, location, temperature and relative humidity as independent variables. Fixed and mobile PM2.5 labels with space-time overlap (i.e. in the same grid at the same hour) are used as response variables. We use a linear regression model to calibrate the mobile data, with goodness of fit $R^2=0.70$ and p-value $<0.05$.

\subsection{Deep-MAPS}
\begin{figure}[h!]
\centering
\includegraphics[width=\textwidth]{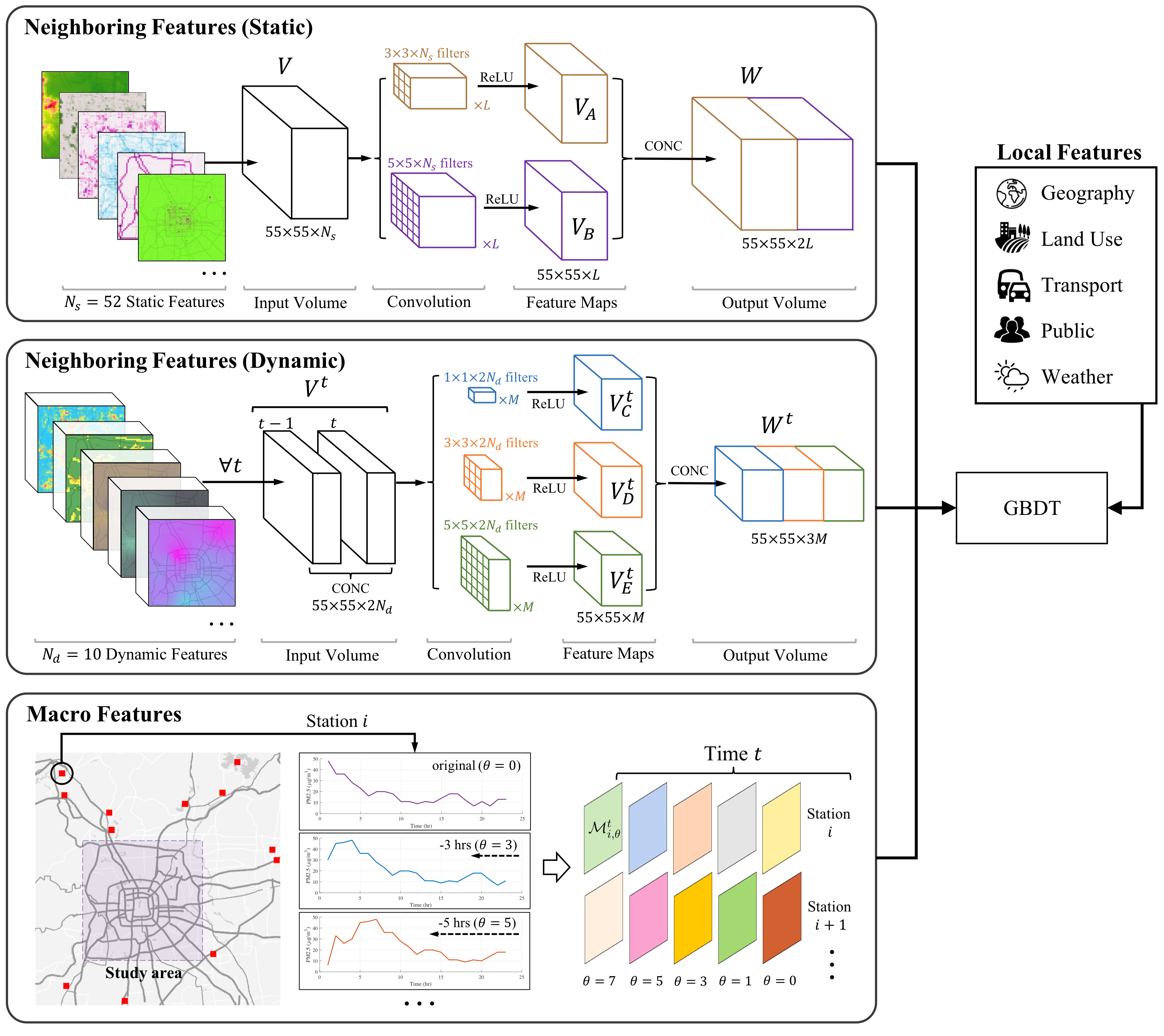}
\caption{\small Feature extraction framework for Deep-MAPS.}
\label{figfeatures}
\end{figure}

\vspace{-0.1 in}
The problem of urban air quality inference based on fixed/mobile sensing involves multi-source and heterogenous datasets that significantly differ in space-time resolution, numerical scale, and veracity. Therefore, it is crucial to define and extract relevant features with appropriate spatial and temporal structures to capture not only internal factors contributing to local AQ (such as transportation, buildings, meteorology), but also indicators of external influence (such as regional transport of pollutants). Deep-MAPS is a machine learning framework, which employs Gradient Boosting Decision Trees to fit air pollution data with local, neighboring and macro features extracted from multi-source urban data, as illustrated in Figure \ref{figfeatures}. While variants of GBDT, such as XGBoost \citep{CG2016} and hybrid boosted decision trees \citep{He2014}, were also considered, they yield similar performance as GBDT. Deep neural networks (such as ConvLSTM) typically require much longer training time and are not particularly robust against the input data structure, which leads to non-convergence and over-fitting. For these reasons, GBDT was selected to perform spatial inference of PM2.5.

\subsubsection{Local features}
\vspace{-0.1 in}
Local features include geographic \& land use characteristics, transport, public vitality, and meteorological conditions, which are defined for each grid and time interval (static features are only defined for grids). The local features do not take into account spatial and temporal correlations among the urban features.  

\subsubsection{Neighboring features}
\vspace{-0.1 in}
Neighboring features are defined for each given grid and time interval, by including features of neighboring grids in space and time. This allows us to capture the spatial-temporal correlations among the urban features. The extraction of neighboring features follows the technique of convolution \citep{Nielsen2015}, and we distinguish between static and dynamic features as follows. 

The $N_s$ static features are treated as $N_s$ input images of size $55\times 55$, forming an input volume $V\in \mathbb{R}^{55\times 55\times N_s}$. Two sets of filters, $\{A^1,\,\ldots,\,A^{L} \}\subset \mathbb{R}^{3\times3\times N_s}$ and $\{B^1,\,\ldots,\,B^{L} \}\subset \mathbb{R}^{5\times5\times N_s}$ are applied to $V$ with stride 1 and padding 0, where the $k$-th channel $A_k^i$ ($B_k^i$) of $A^i$ ($B^i$) is a mean filter with a random multiplicative weight following standard Normal distribution. Convolution with these two sets of filters, followed by the rectifier $f(x)\doteq \max(0,\,x)$ as activation, results in two feature maps
\begin{equation}\label{nfeqn1}
V_{A}=\oplus\left(V_A^1,  \ldots, V_A^{L} \right),  \qquad 
V_{B}=\oplus\left(V_{B}^1,  \ldots,  V_{B}^{L} \right)
\end{equation}
\begin{equation}\label{nfeqn2}
\hbox{where}~~V_{A}^i=f\left(V \ast A^i \right),\quad 
V_{B}^i=f\left(V \ast B^i \right) \qquad 1\leq i \leq L
\end{equation}
Here, $\ast$ is the convolution operator; $\oplus$ is the concatenation operator. The mean filters perform arithmetic average within their receptive fields, while the random weights are used to sample different combinations of the input features. Finally, the output is $W=\oplus(V_A,\,V_B)\in\mathbb{R}^{55\times55\times 2L}$; see Figure \ref{figfeatures}(top).

The $N_d$ dynamic features are convolved in a similar way. For every time $t\in\mathcal{T}$, the input volume $V^t$ consists of the dynamic features of the present and previous time steps to account for the temporal dependencies. We then apply three sets of filters $\{C^1,\,\ldots,\,C^M\}\subset\mathbb{R}^{1\times1\times 2N_d}$, $\{D^1,\,\ldots,\,D^M\}\subset\mathbb{R}^{3\times3\times 2N_d}$, $\{E^1,\,\ldots,\,E^M\}\subset\mathbb{R}^{5\times5\times 2N_d}$, with similar structures as $A^i$ and $B^i$, and obtain 
\begin{equation}\label{nfeqn3}
V_C^t=\oplus\left(V_C^{t,1}, \ldots, V_C^{t,M} \right),~~ V_D^t=\oplus\left(V_D^{t,1}, \ldots, V_D^{t,M} \right), ~~ V_E^t=\oplus\left(V_E^{t,1}, \ldots, V_E^{t,M} \right)
\end{equation}
\begin{equation}\label{nfeqn4}
\hbox{where}~~V_C^{t,i}=f\left(V^t\ast C^i \right),\quad V_D^{t,i}=f\left(V^t\ast D^i \right), \quad V_E^{t,i}=f\left(V^t\ast E^i \right)
\end{equation}
The output for time $t$ is $W^t=\oplus(V_C^t,\,V_D^t,\,V_E^t )\in\mathbb{R}^{55\times55\times 3M}$; see Figure \ref{figfeatures}(middle).

\subsubsection{Macro features}
\vspace{-0.1 in}
In many cities in China and around the globe, severe air pollution may be caused by energy, industrial and agricultural activities in surrounding areas. To account for the regional transport of pollutants without significantly expanding the study area, we utilize PM2.5 data from fixed monitoring stations outside the study area as additional features (coined {\it macro features}). Our premise is that such data are likely to be temporally cross-correlated with the background pollutant concentration in Beijing. By introducing the macro features, we also reduce the risk of overfitting as the AQ inference results are less dependent on locally defined features.

We collect PM2.5 concentration data from 12 fixed monitoring stations outside the study area as shown in Figure \ref{figfeatures}(bottom). By analyzing these time series with time shifts depending on the distance of the stations and average wind speed, we may indirectly account for the transport dynamics of PM2.5. Specifically, $\forall 1\leq i\leq 12$, let $\{y_i(t),\,t\in\mathbb{Z}\}$ be the time series data from the $i$-th station. For a given time $t\in\mathcal{T}$, we define the macro features for all the spatial grids at $t$ to be $\{y_i(t-\theta): 1\leq i\leq 12,~\theta\in\Theta \}$, where $\Theta$ is the set of backward time shifts. $\Theta$ is determined in conjunction with the distance from the study area to the $i$-th station, and the average wind speed in between. Alternatively, the macro features can be viewed as a set of $55\times 55$ images: $\{\mathcal{M}_{i,\theta}^t: 1\leq i\leq 12,~\theta\in\Theta,~t\in\mathcal{T}\}\subset\mathbb{R}^{55\times55}$ where $\mathcal{M}_{i,\theta}^t=y_i(t-\theta)\times \bm 1(55,55)$; see Figure \ref{figfeatures} (bottom).

\end{document}